\pdfoutput=1

\documentclass[11pt]{article}

\usepackage{EMNLP2023}

\usepackage{times}
\usepackage{latexsym}

\usepackage[T1]{fontenc}

\usepackage[utf8]{inputenc}

\usepackage{microtype}

\usepackage{inconsolata}

\usepackage{booktabs}
\usepackage{microtype}
\usepackage{todonotes}
\usepackage{enumitem}
\usepackage{graphicx}
\usepackage{xspace}
\usepackage{xcolor}
\usepackage{amsmath}
\usepackage{multirow}
\usepackage{multicol}
\usepackage{caption}
\usepackage{hyperref}
\usepackage{subcaption}
\usepackage{colortbl}
\usepackage{framed}
\usepackage{comment}
\usepackage{footmisc}
\usepackage{listings}

\newcommand{\myoss}{\textsf{torchdistill}\xspace}

\newcommand\YAMLcolonstyle{\color{red}\mdseries}
\newcommand\YAMLkeystyle{\color{black}\bfseries}
\newcommand\YAMLvaluestyle{\color{teal}\mdseries}
\lstdefinelanguage{yaml}{
    frame=single,
    keywords={true,false,null,y,n},
    keywordstyle=\color{darkgray}\bfseries,
    basicstyle=\YAMLkeystyle\small,
    sensitive=false,
    comment=[l]{!},
    columns=fullflexible,
    morecomment=[s]{/*}{*/},
    emph={!import_call},
    emphstyle=\color{brown},
    commentstyle=\color{brown}\ttfamily\bfseries,
    numberstyle=\color{blue},
    stringstyle=\YAMLvaluestyle,
    moredelim=[l][\color{orange}]{\&},
    moredelim=[l][\color{teal}\mdseries]{*},
    moredelim=**[il][\YAMLcolonstyle{:}\YAMLvaluestyle]{:},
    morestring=[b]',
    morestring=[b]"
}
\definecolor{mygreen}{rgb}{0,0.6,0}
\definecolor{mygray}{rgb}{0.5,0.5,0.5}
\definecolor{mymauve}{rgb}{0.58,0,0.82}
\lstdefinelanguage{mypython}{
    language=Python,
    frame=single,
    basicstyle=\ttfamily\scriptsize,
    comment=[l]{\#},
    keywordstyle=\ttfamily \color{blue},
    stringstyle=\color{teal}
}

\def\y{\mathbf{y}}

\title{\myoss Meets Hugging Face Libraries for Reproducible, Coding-Free Deep Learning Studies: A Case Study on NLP}

\author{Yoshitomo Matsubara~\thanks{~~This work was done prior to joining Amazon.} \\
  University of California, Irvine \\
  \texttt{yoshitom@uci.edu}
}

\begin{document}
\maketitle

\begin{abstract}
Reproducibility in scientific work has been becoming increasingly important in research communities such as machine learning, natural language processing, and computer vision communities due to the rapid development of the research domains supported by recent advances in deep learning.
In this work, we present a significantly upgraded version of \myoss\footnote{\url{https://github.com/yoshitomo-matsubara/torchdistill/} \label{fn:torchdistill}}, a modular-driven coding-free deep learning framework significantly upgraded from the initial release, which supports only image classification and object detection tasks for reproducible knowledge distillation experiments. 
To demonstrate that the upgraded framework can support more tasks with third-party libraries, we reproduce the GLUE benchmark results of BERT models using a script based on the upgraded \myoss, harmonizing with various Hugging Face libraries.
All the 27 fine-tuned BERT models and configurations to reproduce the results are published at Hugging Face\footnote{\url{https://huggingface.co/yoshitomo-matsubara} \label{fn:hf}}, and the model weights have already been widely used in research communities.
We also reimplement popular small-sized models and new knowledge distillation methods and perform additional experiments for computer vision tasks.
\end{abstract}

\section{Introduction}
\label{sec:introduction}

The rapid developments of various research domains such as natural language procession (NLP), computer vision, and speech recognition~\citep{he2016deep,balle2017end,devlin2019bert,dosovitskiy2020image,raffel2020exploring,rombach2022high,radford2023robust} have been supported by advances in deep learning~\citep{krizhevsky2012imagenet,mikolov2013distributed,kingma2014auto,sutskever2014sequence,kingma2015adam,sohl2015deep,vaswani2017attention,brown2020language}.
While it has been developed rapidly, poor reproducibility of deep learning-based studies is a severe problem that research communities have been facing~\citep{crane2018questionable,yang2019critically,daoudi2021lessons,matsubara2021torchdistill}, and the reproducibility has been attracting significant attention from researchers~\citep{gundersen2018on,gundersen2019standing,dodge2019show,kamphuis2020bm25,lopresti2021reproducibility,pineau2021improving}.

To address the serious problem, research communities introduced reproducibility checklists.
At the time of writing, some venues require authors to complete checklists when submitting their work \emph{e.g.}, Responsible NLP Research Checklist\footnote{\url{https://aclrollingreview.org/responsibleNLPresearch/}}~\citep{rogers2021just} at NLP venues (ACL, NAACL, ARR) and Paper Checklist at NeurIPS.\footnote{\url{https://neurips.cc/public/guides/PaperChecklist}}

\begin{figure*}[t]
    \centering
    \includegraphics[width=\linewidth]{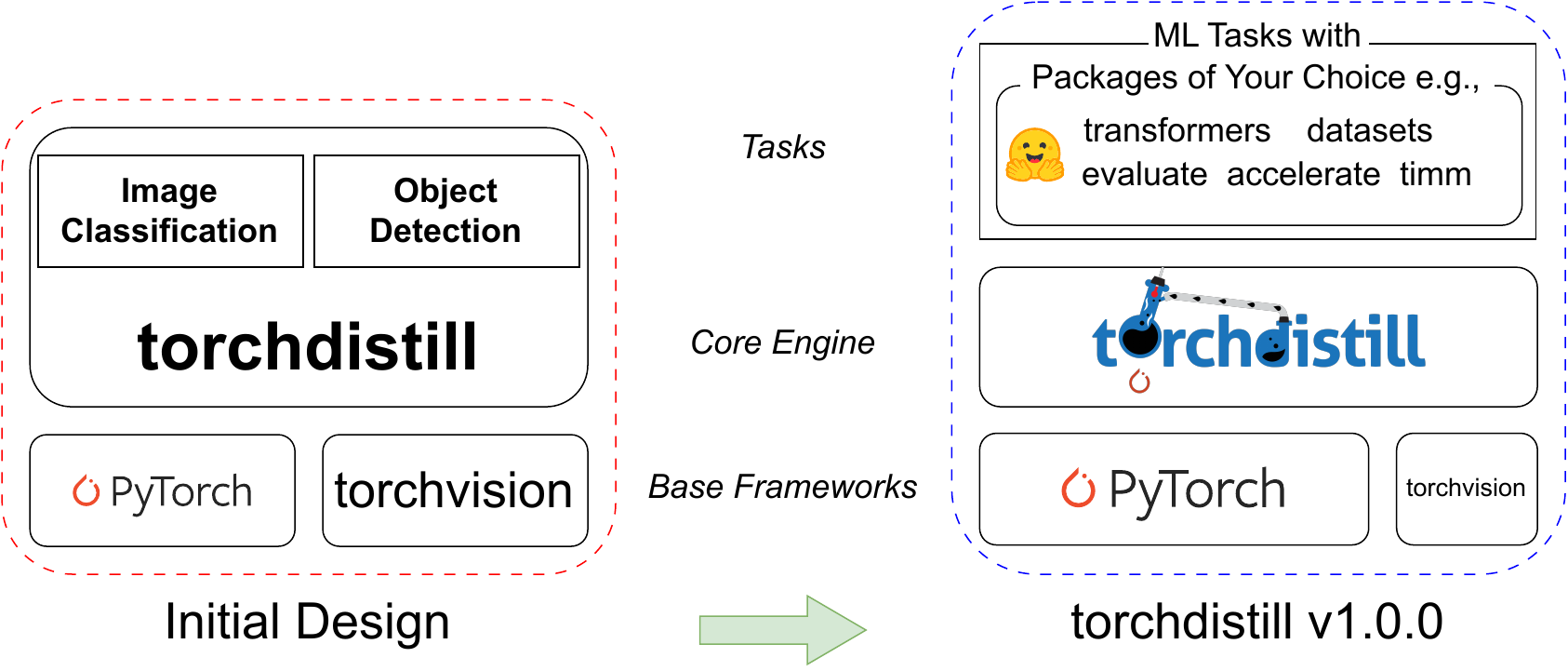}
    \vspace{-0.5em}
    \caption{Initial design of \myoss~\citep{matsubara2021torchdistill} vs. v1.0.0 in this work.}
    \label{fig:arch}
\end{figure*}

\citet{matsubara2021torchdistill} developed \myoss, a modular, configuration-driven knowledge distillation framework built on PyTorch~\citep{paszke2019pytorch} for reproducible deep learning research.
Knowledge distillation~\citep{hinton2014distilling} is a well known model compression method usually to train a small model (called \emph{student}) leveraging outputs from a more complex model (called \emph{teacher}) as part of loss functions to be minimized.
Recent knowledge distillation approaches are more complex \emph{e.g.}, using intermediate layers' outputs (embeddings or feature maps) besides the final output (logits) of teacher models with auxiliary module branches attached to teacher and/or student models during training~\citep{kim2018paraphrasing,zhang2020prime,chen2021distilling}, using multiple teachers~\citep{mirzadeh2020improved,matsubara2022ensemble}, and training multilingual or non-English models solely with an English teacher model~\citep{reimers2020making,li2022learning,gupta2023cross}.

For implementing such approaches, researchers unpacked existing model implementations and modified their input-output interfaces to extract and/or hard-code new auxiliary modules (trainable modules to be used only during training)~\citep{zagoruyko2016wide,passalis2018learning,heo2019knowledge,park2019relational,tian2019contrastive,xu2020knowledge,chen2021distilling}.
\myoss~\citep{matsubara2021torchdistill} was initially designed as a unified knowledge distillation framework to enable users to design experiments by declarative PyYAML configuration files without such hardcoding effort and help researchers complete the ML Code Completeness Checklist\footnote{\url{https://github.com/paperswithcode/releasing-research-code}} for high-quality reproducible knowledge distillation studies.
One of its key concepts is that a declarative PyYAML configuration file designs an experiment and explains key hyperparameters and components used in the experiment.
While the initial framework is well generalized and supports 18 different knowledge distillation methods implemented in a unified way, the implementation of the initial framework is highly dependent on torchvision\footnote{\url{https://github.com/pytorch/vision}}, a package for popular datasets, model architectures, and common image transformations for computer vision tasks.

In this work, we significantly upgrade \myoss from the initial framework~\citep{matsubara2021torchdistill} to enable further generalized implementations, supporting more flexible module abstractions and enhance the advantage of decralative PyYAML configuration files to design experiments with third-party packages of user's choice, as promised in~\citep{matsubara2021torchdistill}.
Using GLUE tasks~\citep{wang2019glue} as an example, we demonstrate that the upgraded \myoss and a new script harmonize with Hugging Face Transformers~\citep{wolf2020transformers}, Datasets~\citep{lhoest2021datasets}, Accelerate~\citep{gugger2022accelerate}, and Evaluate~\citep{von2022evaluate} to reproduce the GLUE test results reported in~\citep{devlin2019bert} by fine-tuning pretrained BERT-Base and BERT-Large models with the upgraded \myoss.
We also conduct knowledge distillation experiments using the fine-tuned BERT-Large models as teachers to train BERT-Base models.
All these experiments are performed on Google Colaboratory.\footnote{\url{https://colab.google/} \label{fn:colab}}
We also publish all the code and configuration files at GitHub\textsuperscript{\ref{fn:torchdistill}} and trained model weights and training logs at Hugging Face\textsuperscript{\ref{fn:hf}} for reproducibility and helping researchers build on this work.
Our BERT models fine-tuned for the GLUE tasks have already been downloaded 138,000 times in total and widely used in research communities not only in research papers but also in tutorials of deep learning frameworks and ACL 2022.
Besides the NLP tasks, we reimplement popular small-sized computer vision models and a few more recent knowledge distillation methods as part of \myoss, and perform additional experiments to demonstrate that the upgraded \myoss still supports computer vision tasks. 

\section{Related Work}
\label{sec:related_work}

In this section, we briefly summarize related work on open source software that supports end-to-end research frameworks.
\citet{yang2018anserini} propose Anserini, an information retrieval toolkit built on Lucene\footnote{\url{https://lucene.apache.org/}} for reproducible information retrieval research.
Pyserini~\citep{lin2021pyserini} is a Python toolkit built on PyTorch~\citep{paszke2019pytorch} and Faiss~\citep{johnson2019billion} for reproducible information retrieval research with sparse and dense representations, and the sparse representation-based retrieval support comes from Lucene via Anserini.

AllenNLP~\citep{gardner2018allennlp} is a toolkit built on PyTorch for research on deep learning methods in NLP and designed to lower barriers to high quality NLP research \emph{e.g.}, useful NLP module abstractions and defining experiments using declarative configuration files.
Highly inspired by AllenNLP,~\citet{matsubara2021torchdistill} design \myoss, a module, configuration-driven framework built on PyTorch for reproducible knowledge distillation studies.
Similar to AllenNLP, \myoss enables users to design experiments by declarative PyYAML configuration files and supports high-level module abstractions.
For image classification and object detection tasks, its generalized starter scripts and configurations help users implement knowledge distillation methods without much coding cost.
\citet{matsubara2021torchdistill} also reimplement 18 knowledge distillation methods with \myoss and point out that the standard knowledge distillation~\citep{hinton2014distilling} can outperform many of the recent state of the art knowledge distillation methods for a popular teacher-student pair (ResNet-34 and ResNet-18) with ILSVRC 2012 dataset~\citep{russakovsky2015imagenet}.
In Section~\ref{sec:upgrades}, we describe major upgrades in \myoss from the initial release~\citep{matsubara2021torchdistill}.

\section{Major Upgrades from the Initial Release}
\label{sec:upgrades}

\begin{figure*}[!t]
    \centering
    \noindent
    \begin{minipage}[t]{\linewidth}
        \begin{lstlisting}[language=mypython, emph={build_transform}, emphstyle=\color{violet}, backgroundcolor=\color{pink!25}]
import torchvision
from torchdistill.datasets.transform import TRANSFORM_CLASS_DICT

TRANSFORM_CLASS_DICT.update(torchvision.transforms.__dict__)


def build_transform(transform_params_config, compose_cls=None):
    if not isinstance(transform_params_config, (dict, list)) or len(transform_params_config) == 0:
        return None

    component_list = list()
    if isinstance(transform_params_config, dict):
        for component_key in sorted(transform_params_config.keys()):
            component_config = transform_params_config[component_key]
            params_config = component_config.get('params', dict())
            if params_config is None:
                params_config = dict()

            component = TRANSFORM_CLASS_DICT[component_config['type']](**params_config)
            component_list.append(component)
    else:
        for component_config in transform_params_config:
            params_config = component_config.get('params', dict())
            if params_config is None:
                params_config = dict()

            component = TRANSFORM_CLASS_DICT[component_config['type']](**params_config)
            component_list.append(component)
    return transforms.Compose(component_list) if compose_cls is None else compose_cls(component_list)
        \end{lstlisting}
    \end{minipage}
    \noindent
    \begin{minipage}[t]{0.44\linewidth}
    \begin{lstlisting}[language=yaml, backgroundcolor=\color{pink!25}]
transform_params:
  - type: 'RandomCrop'
    params:
      size: 32
      padding: 4
  - type: 'RandomHorizontalFlip'
    params:
      p: 0.5
  - type: 'ToTensor'
    params:
  - type: 'Normalize'
    params:
      mean: [0.49139968, 0.48215841, 0.44653091]
      std: [0.24703223, 0.24348513, 0.26158784]
    \end{lstlisting}
    \end{minipage}\hfill
    \begin{minipage}[t]{0.52\linewidth}
    \begin{lstlisting}[language=yaml, backgroundcolor=\color{cyan!15}]
transform: !import_call
  key: 'torchvision.transforms.Compose'
  init:
    kwargs:
      transforms:
        - !import_call
          key: 'torchvision.transforms.RandomCrop'
          init:
            kwargs:
              size: 32
              padding: 4
        - !import_call
          key: 'torchvision.transforms.RandomHorizontalFlip'
          init:
            kwargs:
              p: 0.5
        - !import_call
          key: 'torchvision.transforms.ToTensor'
          init:
        - !import_call
          key: 'torchvision.transforms.Normalize'
          init:
            kwargs:
              mean: [0.49139968, 0.48215841, 0.44653091]
              std: [0.24703223, 0.24348513, 0.26158784]
    \end{lstlisting}
    \end{minipage}
    \vspace{-1em}
    \caption{Example of two different ways to build a sequence of transforms in torchvision (\textbf{transform}) for CIFAR-10 dataset. The initial version ({\color{pink}top, left}) defines a function for torchvision {\ttfamily\color{violet}build\_transform} in \myoss and gives the function a list of dict objects in the left PyYAML as {\ttfamily transform\_params\_config}. \myoss in this work ({\color{cyan}right}) can build exactly the same \textbf{transform} by instantiating each of the transform classes step-by-step with {\bf\ttfamily\color{brown}!import\_call}, one of our pre-defined PyYAML constructors in the upgraded \myoss.}
    \label{fig:instantiations}
\end{figure*}

In this section, we summarize the major upgrades from the initial release of \myoss~\citep{matsubara2021torchdistill}.
Figure~\ref{fig:arch} highlights high-level differences between the initial design~\citep{matsubara2021torchdistill} of \myoss and a largely upgraded version in this work.
The initial \myoss is dependent on PyTorch and torchvision and contains key modules and functionalities specifically designed to support image classification and object detection tasks.
For example, dataset modules that the initial version officially supports are only those in torchvision, and some of dataset-relevant functionalities such as building a sequence of data transforms and dataset loader are based on datasets in torchvision.

In this work, we make \myoss less dependent on torchvision and support more tasks with third-party packages of users' choice, by generalizing some of the key components in the framework and exporting task-specific implementations to the corresponding executable scripts and local packages.
We also reimplement popular small-sized models whose official PyTorch implementations are not either available or maintained.

\subsection{PyYAML-based Instantiation}
\label{subsec:instantiation}

A declarative PyYAML configuration file plays an important role in \myoss.
Users can design experiments with the declarative PyYAML configuration file, which defines various types of abstracted modules with hyperparameters such as dataset, model, optimizer, scheduler, and loss modules.
To allow more flexibility in PyYAML configurations, we add more useful constructors such as importing arbitrary local packages to register modules but without edits on an executable script, and instantiating an arbitrary class with a log message.
Those can be done simply at the very beginning of an experiment when loading the PyYAML configuration file and make the configuration files more self-explanatory since the configuration format used for the initial version does not explicitly tell users whether the experiment needs specific local packages.
Those features also help us generalize ways to define key module such as datasets and their components (\emph{e.g.}, pre-processing transforms, samplers).

Figure~\ref{fig:instantiations} shows an example that build a sequence of image/tensor transforms with the initial version and \myoss in this work.
While the former requires both a Python function specifically designed for torchvision modules ({\ttfamily\color{violet}build\_transform}) and a list of dict objects defined in a PyYAML configuration to be given to the function as ({\ttfamily transform\_params\_config}), the latter can build exactly the same transform when loading the PyYAML configuration and store the instantiated object as part of a dict object with {\bf\ttfamily transform} key.

\subsection{Generalized Modules for Supporting More Tasks}
\label{subsec:generalized_modules}
The PyYAML-based instantiation feature described in Section~\ref{subsec:instantiation} enables us to remove torchvision-specific modules mentioned in Section~\ref{sec:upgrades} (\emph{e.g.}, {\ttfamily\color{violet}build\_transform} in Fig.~\ref{fig:instantiations}) so that we can reduce \myoss's dependency on torchvision and generalize its modules for supporting more tasks.

The initial version of \myoss is designed to support image classification and object detection tasks based on torchvision, and torchvision models for the tasks such as ResNet~\citep{he2016deep} and Faster R-CNN~\citep{ren2015faster} require an image (tensor) and an annotation as part of the model inputs during training.
However, this interface does not generalize well to support other tasks.
Taking a text classification task as an example, Transformer~\citep{vaswani2017attention} models in Hugging Face Transformers~\citep{wolf2020transformers} have much more input data fields  such as (not limited to) token IDs, attention mask, token type IDs, position IDs, and labels for BERT~\citep{devlin2019bert}, and different models have different input data fields \emph{e.g.}, BART~\citep{lewis2020bart} has additional input data fields such as token IDs for its decoder.

In order to support diverse models and tasks, we generalize interfaces of model input/output and the subsequent processes in \myoss such as computing training losses.
For demonstrating that the upgraded \myoss can support more tasks, we provide starter scripts based on the upgraded framework for GLUE~\citep{wang2019glue} and semantic segmentation tasks.
For the GLUE tasks, we harmonize popular Python libraries with \myoss in the script such as Hugging Face Transformers~\citep{wolf2020transformers}, Datasets~\citep{lhoest2021datasets}, and Evaluate~\citep{von2022evaluate} for model, dataset, and evaluation modules.
We also leverage Accelerate~\citep{gugger2022accelerate} for efficient training and inference.
In Section~\ref{subsec:glue_demos}, we demonstrate GLUE experiments with \myoss and the third-party libraries.

\subsection{Reimplemented Models and Methods}
We find in recent knowledge distillation studies~\citep{tian2019contrastive,xu2020knowledge,chen2021distilling} that there is still a demand of small models for relatively simple datasets such as ResNet~\citep{he2016deep}\footnote{\url{https://github.com/facebookarchive/fb.resnet.torch}}, WRN~\citep{zagoruyko2016wide}\footnote{\url{https://github.com/szagoruyko/wide-residual-networks}}, and DenseNet~\citep{huang2017densely}\footnote{\url{https://github.com/liuzhuang13/DenseNet}} for image classification tasks with CIFAR-10 and CIFAR-100 datasets~\citep{krizhevsky2009learning} since the official repositories are no longer maintained and/or not implemented with PyTorch.

For helping the community conduct better benchmarking, we reimplement the models for CIFAR-10 and CIFAR-100 datasets as part of \myoss and attempt to reproduce the reported results following the original training recipes (See Section~\ref{sec:demos}).
With the upgraded \myoss, we also reimplement and test a few more knowledge distillation methods~\citep{he2019knowledge,chen2021distilling}.

\section{Google Colab Demos}
\label{sec:demos}

\begin{table*}[t]
    \def\arraystretch{1.1}
    \begin{center}
        \bgroup
        \setlength{\tabcolsep}{0.25em}
        \small
        \begin{tabular}{l|rrrrrrrrr}
            \multicolumn{1}{c|}{\multirow{2}{*}{\bf Model (Method, Reference)}} & \multicolumn{1}{c}{\bf MNLI-(m/mm)} & \multicolumn{1}{c}{\bf QQP} & \multicolumn{1}{c}{\bf QNLI} & \multicolumn{1}{c}{\bf SST-2} & \multicolumn{1}{c}{\bf CoLA} & \multicolumn{1}{c}{\bf STS-B} & \multicolumn{1}{c}{\bf MRPC} & \multicolumn{1}{c}{\bf RTE} & \multicolumn{1}{c}{\bf WNLI} \\
            & \textbf{Acc./Acc.} & \textbf{F1} & \textbf{Acc.} & \textbf{Acc.} & \textbf{M Corr.} & \textbf{P-S Corr.} & \textbf{F1} & \textbf{Acc.} & \textbf{Acc.} \\
            \midrule
            BERT-Large (FT,~\citet{devlin2019bert}) & 86.7/85.9 & 72.1 & 92.7 & 94.9 & 60.5 & 86.5 & 89.3 & 70.1 & N/A \\
            BERT-Large (FT, Ours) & \href{https://huggingface.co/yoshitomo-matsubara/bert-large-uncased-mnli}{86.4/85.7} & \href{https://huggingface.co/yoshitomo-matsubara/bert-large-uncased-qqp}{72.2} & \href{https://huggingface.co/yoshitomo-matsubara/bert-large-uncased-qnli}{92.4} & \href{https://huggingface.co/yoshitomo-matsubara/bert-large-uncased-sst2}{94.6} & \href{https://huggingface.co/yoshitomo-matsubara/bert-large-uncased-cola}{61.5} & \href{https://huggingface.co/yoshitomo-matsubara/bert-large-uncased-stsb}{85.0} & \href{https://huggingface.co/yoshitomo-matsubara/bert-large-uncased-mrpc}{89.2} & \href{https://huggingface.co/yoshitomo-matsubara/bert-large-uncased-rte}{68.9} & \href{https://huggingface.co/yoshitomo-matsubara/bert-large-uncased-wnli}{65.1} \\
            \midrule
            BERT-Base (FT,~\citet{devlin2019bert}) & 84.6/83.4 & 71.2 & 90.5 & 93.5 & 52.1 & 85.8 & 88.9 & 66.4 & N/A \\
            BERT-Base (FT, Ours) & \href{https://huggingface.co/yoshitomo-matsubara/bert-base-uncased-mnli}{84.2/83.3} & \href{https://huggingface.co/yoshitomo-matsubara/bert-base-uncased-qqp}{71.4} & \href{https://huggingface.co/yoshitomo-matsubara/bert-base-uncased-qnli}{91.0} & \href{https://huggingface.co/yoshitomo-matsubara/bert-base-uncased-sst2}{94.1} & \href{https://huggingface.co/yoshitomo-matsubara/bert-base-uncased-cola}{51.1} & \href{https://huggingface.co/yoshitomo-matsubara/bert-base-uncased-stsb}{84.4}
            & \href{https://huggingface.co/yoshitomo-matsubara/bert-base-uncased-mrpc}{86.8} & \href{https://huggingface.co/yoshitomo-matsubara/bert-base-uncased-rte}{66.7} & \href{https://huggingface.co/yoshitomo-matsubara/bert-base-uncased-wnli}{65.8} \\
            BERT-Base (KD, Ours) & \href{https://huggingface.co/yoshitomo-matsubara/bert-base-uncased-mnli_from_bert-large-uncased-mnli}{85.9/84.7} & \href{https://huggingface.co/yoshitomo-matsubara/bert-base-uncased-qqp_from_bert-large-uncased-qqp}{72.8} & \href{https://huggingface.co/yoshitomo-matsubara/bert-base-uncased-qnli_from_bert-large-uncased-qnli}{90.7} & \href{https://huggingface.co/yoshitomo-matsubara/bert-base-uncased-sst2_from_bert-large-uncased-sst2}{93.7} & \href{https://huggingface.co/yoshitomo-matsubara/bert-base-uncased-cola_from_bert-large-uncased-cola}{57.0} & \href{https://huggingface.co/yoshitomo-matsubara/bert-base-uncased-stsb_from_bert-large-uncased-stsb}{85.6} & \href{https://huggingface.co/yoshitomo-matsubara/bert-base-uncased-mrpc_from_bert-large-uncased-mrpc}{87.5} & \href{https://huggingface.co/yoshitomo-matsubara/bert-base-uncased-rte_from_bert-large-uncased-rte}{66.7} & \href{https://huggingface.co/yoshitomo-matsubara/bert-base-uncased-wnli_from_bert-large-uncased-wnli}{65.1} \\
        \end{tabular}
        \egroup
    \end{center}
    \vspace{-0.5em}
    \caption{GLUE test results. Our results are hyperlinked to our Hugging Face Model repositories. FT: Fine-Tuning, KD: Knowledge Distillation using BERT-Large (FT, ours) as teacher.}
    \label{table:glue_results}
\end{table*}

In this section, we demonstrate that the upgraded \myoss can collaborate with third-party libraries for supporting more tasks.
We also attempt to reproduce the CIFAR-10 and CIFAR-100 results reported in the original papers.
To lower the barrier to reusing and building on the scripts with \myoss, we conduct all the experiments on Google Colaboratory\textsuperscript{\ref{fn:colab}}, which gives users access to GPUs free of charge.
We publish the Jupyter Notebook\footnote{\url{https://jupyter.org/}} files to run the experiments as part of \myoss repository\textsuperscript{\ref{fn:torchdistill}} so that researchers can easily use them.

\subsection{GLUE Tasks}
\label{subsec:glue_demos}

The GLUE benchmark~\citep{wang2019glue} uses nine datasets in three different task categories.
The benchmark consists of 1) two single-sentence tasks: CoLA~\citep{warstadt2019neural} and SST-2~\citep{socher2013recursive}, 2) three similarity and paraphrase tasks: MRPC~\citep{dolan2005automatically}, QQP\footnote{\url{https://quoradata.quora.com/First-Quora-Dataset-Release-Question-Pairs}}, and STS-B~\citep{cer2017semeval}, and 3) four inference tasks: MNLI~\citep{williams2018broad}, QNLI~\citep{rajpurkar2016squad,wang2019glue}, RTE~\citep{dagan2005pascal,haim2006second,giampiccolo2007third,bentivoglisixth}, and WNLI~\citep{levesque2012winograd}.

We attempt to reproduce GLUE test results reported in a popular study, BERT~\citep{devlin2019bert}, using the upgraded \myoss harmonizing with Hugging Face libraries (transformers, datasets, evaluate, and accelerate)~\citep{wolf2020transformers,lhoest2021datasets,von2022evaluate,gugger2022accelerate}.
Following the experiments, we also conduct knowledge distillation experiments that fine-tune pretrained BERT-Base models for GLUE tasks, using the fine-tuned BERT-Large models as teachers for the knowledge distillation method of~\citet{hinton2014distilling} minimizing
\begin{equation}
    \mathcal{L} = \alpha \cdot \mathcal{L}_\text{CE}(\mathbf{\hat{y}}, \mathbf{y}) + (1 - \alpha) \cdot \tau^2 \cdot \mathcal{L}_\text{KL} \left(\mathbf{p}, \mathbf{q}\right),
    \label{eq:kd}
\end{equation}
\noindent where $\mathcal{L}_\text{CE}$ is a standard cross entropy.
$\hat{\y}$ indicates the student model's estimated class probabilities, and $\y$ is the annotated category.
$\mathcal{L}_\text{KL}$ is the Kullback-Leibler divergence, and $\alpha$ and $\tau$ are a balancing factor and a temperature, respectively.
$\mathbf{p}$ and $\mathbf{q}$ represent the \emph{softened} output distributions from teacher and student models, respectively.
$\mathbf{p}$ is used as a target distribution for $\mathcal{L}_\text{KL}$.
Specifically, $\mathbf{p} = [p_1, p_2, \ldots, p_{|\mathcal{C}|}]$ where $\mathcal{C}$ is a set of categories in the target task.
$p_i$ indicates the student model's softened output value (scalar) for the $i$-th category:
\begin{equation}
    p_{i} = \frac{\exp \left( \frac{v_i}{\tau} \right)}{\sum_{k \in \mathcal{C}} \exp\left( \frac{v_k}{\tau} \right)},
\end{equation}
\noindent where $\tau$ is one of the hyperparameters defined in Eq. (\ref{eq:kd}).
$v_i$ denotes a logit value for the $i$-th category. 
The same rules are applied to $\mathbf{q}$ for the student model.

For reproducing the GLUE test results in~\citep{devlin2019bert}, we use pretrained BERT-Base\footnote{\url{https://huggingface.co/bert-base-uncased}} and BERT-Large\footnote{\url{https://huggingface.co/bert-large-uncased}} models in Hugging Face Transformers~\citep{wolf2020transformers}.
Following~\citep{devlin2019bert} we minimize a standard cross-entropy and the Adam optimizer~\citep{kingma2015adam} with slightly extended hyperparameter choices: batch size of either 16 or 32 and 2-5 epochs for fine-tuning and select a learning rate among $\{2.0 \times 10^{-5}, 3.0 \times 10^{-5}, 4.0 \times 10^{-5}, 5.0 \times 10^{-5}\}$ on the \emph{dev} set for each of the tasks.
For knowledge distillation, we also choose learning rate from $\{1.0 \times 10^{-5}, 2.0 \times 10^{-5}, 3.0 \times 10^{-5}, 4.0 \times 10^{-5}, 5.0 \times 10^{-5}\}$, temperature $\tau \in \{1, 3, 5, 7, 9, 11\}$, and a balancing weight $\alpha \in \{0.1, 0.3, 0.5, 0.7, 0.9\}$ based on the \emph{dev} sets.
Note that since STS-B is not a classification task, we use the sum of 1) a mean squared error between the annotation and the student model's output and 2) a mean squared error between outputs of the teacher and student models instead of Eq. (\ref{eq:kd}) for the dataset.

Table~\ref{table:glue_results} shows the GLUE test results reported by~\citet{devlin2019bert} and those obtained from GLUE Benchmark\footnote{\url{https://gluebenchmark.com/}} for our three configurations: fine-tuning pretrained BERT-Base (FT, Ours) and pretrained BERT-Large (FT, Ours) models and knowledge distillation to fine-tune pretrained BERT-Base (KD, Ours) as a student, using the fine-tuned BERT-Large as the teacher.
Note that~\citet{devlin2019bert} do not report the results for the WNLI test dataset.

Overall, our fine-tuned BERT-Base and BERT-Large models achieved GLUE test results comparable to the official test results reported by~\citet{devlin2019bert}.
The knowledge distillation method~\citep{hinton2014distilling} helped BERT-Base models improve the performance for most of the tasks, compared to those fine-tuned without the teacher models.
All the trained model weights and training logs are published at Hugging Face\textsuperscript{\ref{fn:hf}}, and the training configurations are published as part of the \myoss GitHub repository.\textsuperscript{\ref{fn:torchdistill}}

The fine-tuned BERT models we published are widely used in the research communities and have already been downloaded about 138,000 times in total at the time of writing.
For instance, some of the models are used for benchmarks, ensembling, model quantization, token pruning~\citep{matena2022merging,church2022gentle,guo2022ant,lee2022sparse}, DeepSpeed Tutorials\footnote{\url{https://www.deepspeed.ai/tutorials/model-compression/}}, Intel\textregistered~Neural Compressor Examples\footnote{\url{https://github.com/intel/neural-compressor/tree/master/examples}}, and ACL 2022 Tutorial.\footnote{\url{https://github.com/kwchurch/ACL2022_deepnets_tutorial}}

\subsection{CIFAR-10 and CIFAR-100}
\label{subsec:cifar_demos}

We also attempt to reproduce the CIFAR-10 and CIFAR-100 results reported in~\citep{he2016deep,zagoruyko2016wide,huang2017densely} using the upgraded \myoss with the reimplemented ResNet, WRN, and DenseNet models.
We follow the original papers and reuse the hyperparameter choices and training recipes such as data augmentations.
Note that we do not confider models that can not fit to the GPU memory which Google Colab can offer \emph{e.g.}, ResNet-1202~\citep{he2016deep} for CIFAR-10 and DenseNet-BC($k=24$ and $k=40$)~\citep{huang2017densely} for CIFAR-10 and CIFAR-100.

\begin{table}[t]
    \def\arraystretch{1.1}
    \begin{center}
        \bgroup
        \setlength{\tabcolsep}{0.35em}
        \small
        \begin{tabular}{l|rr}
            \multicolumn{1}{c|}{\multirow{2}{*}{\bf CIFAR-10 Model}} & \multicolumn{2}{c}{\bf Test Accuracy} \\
            & \textbf{Original} & \textbf{\myoss} \\
            \midrule
            ResNet-20 & 91.25 & 91.92 \\
            ResNet-32 & 92.49 & 93.03 \\
            ResNet-44 & 92.83 & 93.20 \\
            ResNet-56 & 93.03 & 93.57 \\
            ResNet-110 & 93.57 & 93.50 \\
            WRN-40-4 & 95.47 & 95.24 \\
            WRN-28-10 & 96.00 & 95.53 \\
            WRN-16-8 & 95.73 & 94.76 \\
            DenseNet-BC (k=12, depth=100) & 95.49 & 95.53 \\
        \end{tabular}
        \egroup
    \end{center}
    \vspace{-0.5em}
    \caption{CIFAR-10 results for ResNet~\citep{he2016deep}, WRN~\citep{zagoruyko2016wide}, and DenseNet~\citep{huang2017densely}.}
    \label{table:cifar10_ce_results}
    \vspace{0.5em}
    \begin{center}
        \bgroup
        \setlength{\tabcolsep}{0.35em}
        \small
        \begin{tabular}{l|rr}
            \multicolumn{1}{c|}{\multirow{2}{*}{\bf CIFAR-100 Model}} & \multicolumn{2}{c}{\bf Test Accuracy} \\
            & \textbf{Original} & \textbf{\myoss} \\
            \midrule
            WRN-40-4 & 79.82 & 79.44 \\
            WRN-28-10 & 80.75 & 81.27 \\
            WRN-16-8 & 79.57 & 79.26 \\
            DenseNet-BC (k=12, depth=100) & 77.73 & 77.14 \\
        \end{tabular}
        \egroup
    \end{center}
    \vspace{-0.5em}
    \caption{CIFAR-100 results for WRN~\citep{zagoruyko2016wide} and DenseNet~\citep{huang2017densely}.}
    \label{table:cifar100_ce_results}
\end{table}

Tables~\ref{table:cifar10_ce_results} and~\ref{table:cifar100_ce_results} compare the results reported in the original papers~\citep{he2016deep,zagoruyko2016wide,huang2017densely} with those we reproduced for CIFAR-10 and CIFAR-100 test datasets, respectively.
We can confirm that for most of the reimplemented models, our results are comparable to those reported in the original papers.
Those model weights and training configuration files are publicly available, and users can automatically download the weights via the upgraded \myoss PyPI package.

\section{ILSVRC 2012}
\label{sec:imagenet}

As highlighted in Section~\ref{sec:upgrades}, \myoss was initially focused on supporting implementations of diverse knowledge distillation in a unified way and dependent on torchvision to specifically support image classification and object detection tasks with its relevant modules (see Fig.~\ref{fig:arch}).
To demonstrate that the upgraded \myoss still preserves the feature, we reimplement a few more knowledge distillation methods with the upgraded \myoss: knowledge review (KR) framework~\citep{chen2021distilling} and knowledge translation and adaptation with affinity distillation  (KTAAD)~\citep{he2019knowledge}.
Note that~\citet{matsubara2021torchdistill} present the results of various knowledge distillation methods reimplemented with the initial version of \myoss for ILSVRC 2012 and COCO 2017~\citep{lin2014microsoft} datasets.
Those results are not included in this work, and we refer interested readers to~\citep{matsubara2021torchdistill}.

\citet{chen2021distilling} demonstrate that the KR method can outperform other knowledge distillation using ResNet-34 and ResNet-18~\citep{he2016deep}, a popular pair of teacher and student models for the ImageNet (ILSVRC 2012) dataset~\citep{russakovsky2015imagenet}.
Using the reimplemented KR method based on the upgraded \myoss with hyperparameters in~\citep{chen2021distilling}, we successfully reproduce their reported result of ResNet-18 for the ImageNet dataset as shown in Table~\ref{table:kr_results}.
The trained model weights and configuration are published as part of the \myoss repository.\textsuperscript{\ref{fn:torchdistill}}

\begin{table}[tb]
    \begin{center}
        \bgroup
        \setlength{\tabcolsep}{0.3em}
        \small
        \begin{tabular}{r|rrr}
            \multicolumn{1}{c|}{\bf T: ResNet-34} & \multicolumn{3}{c}{\bf S: ResNet-18} \\
            \multicolumn{1}{c|}{\bf CE} & \multicolumn{1}{c}{\bf CE} & \multicolumn{1}{c}{\bf KR (Original)} & \multicolumn{1}{c}{\bf KR (Ours)}\\
            \midrule
            73.31 & 69.75 & 71.61 & 71.64 \\
        \end{tabular}
        \egroup
    \end{center}
    \caption{ILSVRC 2012 top-1 accuracy of ResNet-18 (student) trained by KR~\citep{chen2021distilling} with pretrained ResNet-34 (teacher). CE: torchvision models pretrained with cross-entropy.}
    \label{table:kr_results}
\end{table}

\section{PASCAL VOC 2012 \& COCO 2017}
\label{sec:coco_pascal_voc}
The initial \myoss~\citep{matsubara2021torchdistill} supports image classification and object detection tasks.
As mentioned in Section~\ref{subsec:generalized_modules}, we also provide a starter script for semantic segmentation tasks.
Using two popular datasets, PASCAL VOC 2012~\citep{everingham2012pascal} and COCO 2017~\citep{lin2014microsoft}, we demonstrate that the upgraded \myoss supports semantic segmentation tasks as well.

In the experiments with PASCAL VOC 2012 dataset, we use DeepLabv3~\citep{chen2017rethinking} with ResNet-50 and ResNet-101 backbones~\citep{he2016deep}, using torchvision's pretrained model weights for COCO 2017 dataset.
We choose hyperparameters such as learning rate policy and crop size based on the original study of DeepLabv3~\citep{chen2017rethinking}.
Our results are shown in Table~\ref{table:pascal_voc_results}, and DeepLabv3 with ResNet-101 achieved comparable mIoU (mean Intersection over Union) to the best DeepLabv3 model for PASCAL VOC 2012 dataset (\emph{val} set) in the original study (mIoU: 82.70).
Following torchvision documentation\footnote{\url{https://pytorch.org/vision/stable/models.html\#table-of-all-available-semantic-segmentation-weights}}, we measure global pixelwise accuracy as well.
In terms of both the metrics, DeepLabv3 with ResNet-101 outperforms DeepLabv3 with ResNet-50.

We also examine our reimplemented KTAAD method~\citep{he2019knowledge} for the Lite R-ASPP model (LRASPP in torchvision)~\citep{howard2019searching} as a student model, using the COCO 2017 dataset and the pretrained DeepLabv3 with ResNet-50 in torchvision as a teacher model, whose mIoU and global pixelwise accuracy are 66.4 and 92.4, respectively.
Since the KTAAD method is not tested on COCO 2017 dataset for LRASPP with MobileNetV3-Large backbone in the original paper of KTAAD~\citep{he2019knowledge}, our hyperparameter choice is based on torchvision's reference script.\footnote{\url{https://github.com/pytorch/vision/tree/main/references/segmentation}}

Table~\ref{table:coco_results} presents the semantic segmentation results of LRASPP with MobileNetV3-Large backbone trained without the teacher model and by the KTAAD method we reimplemented.
We confirm that the student model trained by KTAAD outperforms the same model trained on COCO 2017 available in torchvision in terms of mean IoU and global pixelwise accuracy.

As with other experiments, the trained model weights and configuration used in this section are published as part of the \myoss repository.\textsuperscript{\ref{fn:torchdistill}}

\begin{table}[tb]
    \begin{center}
        \bgroup
        \setlength{\tabcolsep}{0.3em}
        \small
        \begin{tabular}{l|rr}
            \multicolumn{1}{c|}{\bf Model} & \multicolumn{1}{c}{\bf mean IoU} & \multicolumn{1}{c}{\bf Pixelwise Acc.} \\
            \midrule
            DeepLabv3 w/ ResNet-50 & 80.6 & 95.7 \\
            DeepLabv3 w/ ResNet-101 & 82.4 & 96.2 \\
        \end{tabular}
        \egroup
    \end{center}
    \vspace{-0.5em}
    \caption{PASCAL VOC 2012 (Segmentation, \emph{val} set) validation results for DeepLabv3 with ResNet backbones~\citep{chen2017rethinking} initialized with torchvision pretrained model weights for COCO 2017 dataset.}
    \label{table:pascal_voc_results}
    \vspace{1.0em}
    \begin{center}
        \small
        \begin{tabular}{l|rr}
            \multicolumn{1}{c|}{\bf Method} & \multicolumn{1}{c}{\bf mean IoU} & \multicolumn{1}{c}{\bf Pixelwise Acc.} \\
            \midrule
            CE (torchvision) & 57.9 & 91.2 \\
            KTAAD (Ours) & 58.2 & 92.1 \\
        \end{tabular}
    \end{center}
    \vspace{-0.5em}
    \caption{COCO 2017 (Segmentation, \emph{val} set) results for LRASPP with MobileNetV3-Large backbone~\citep{howard2019searching}.}
    \label{table:coco_results}
\end{table}

\section{Conclusion}
In this work, we significantly upgraded \myoss~\citep{matsubara2021torchdistill}, a modular, configuration-driven framework built on PyTorch~\citep{paszke2019pytorch} for reproducible deep learning and knowledge distillation studies.
We enhanced PyYAML-based instantiation, generalized internal modules for supporting more tasks, and reimplemented popular models and methods.

To demonstrate that the upgraded framework can support more tasks as we claim, we provided starter scripts for new tasks based on the upgraded framework.
One of the new starter scripts supports GLUE tasks~\citep{wang2019glue} and harmonizes with Hugging Face Transformers~\citep{wolf2020transformers}, Datasets~\citep{lhoest2021datasets}, Accelerate~\citep{gugger2022accelerate}, and Evaluate~\citep{von2022evaluate}. 
Using the script on Google Colaboratory, we reproduced the GLUE test results of fine-tuned BERT models~\citep{devlin2019bert} and performed  knowledge distillation experiments with our fine-tuned BERT-Large models as teacher models.
Similarly, we reproduced CIFAR-10 and -100 results of popular small-sized models we reimplemented, using Google Colaboratory.
Furthermore, we reproduced the result of ResNet-18 trained with the reimplemented KR method~\citep{chen2021distilling} for the ImageNet dataset.
We also demonstrated a new starter script for semantic segmentation tasks using PASCAL VOC 2012 and COCO 2017 datasets, and the reimplemented KTAAD method~\citep{he2019knowledge} improves a pretrained semantic segmentation model in torchvision.

In this study, we also published 27 trained models for NLP tasks\textsuperscript{\ref{fn:hf}} and 14 trained models for computer vision tasks.\textsuperscript{\ref{fn:torchdistill}} 
According to Hugging Face Model repositories, the BERT models fine-tuned for the GLUE tasks have already been downloaded about 138,000 times in total at the time of writing.
Research communities leverage \myoss not only for knowledge distillation studies~\citep{liu2021exploring,li2022role,lin2022knowledge,dong2022soteacher,miles2023closer}, but also for machine learning reproducibility challenge (MLRC)~\citep{lee2023re} and reproducible deep learning studies~\citep{matsubara2022bottlefit,matsubara2022supervised,furutanpey2023frankensplit,furutanpey2023architectural,matsubara2023sc2}.
\myoss is publicly available as a pip-installable PyPI package and will be maintained and upgraded for encouraging coding-free reproducible deep learning and knowledge distillation studies.

\section*{Acknowledgements}
We thank the anonymous reviewers for their comments. 
This project has been supported by Travis CI's OSS credits and JetBrain's Free License Programs (Open Source) since November 2021 and June 2022, respectively.

\bibliography{references}
\bibliographystyle{acl_natbib}

\end{document}